\documentclass{article}

\PassOptionsToPackage{numbers, compress}{natbib}
\usepackage[preprint]{nips_2018}

\usepackage[utf8]{inputenc} % allow utf-8 input
\usepackage[T1]{fontenc}    % use 8-bit T1 fonts
\usepackage{hyperref}       % hyperlinks
\usepackage{url}            % simple URL typesetting
\usepackage{booktabs}       % professional-quality tables
\usepackage{amsfonts}       % blackboard math symbols
\usepackage{nicefrac}       % compact symbols for 1/2, etc.
\usepackage{microtype}      % microtypography

\usepackage{graphicx}
\usepackage{subfigure}

\usepackage{algorithm}
\usepackage{algorithmicx}
\usepackage{algpseudocode}
\usepackage{amsmath}

\title{Variational Capsules for Image Analysis and Synthesis}

\author{
    Huaibo Huang, Lingxiao Song, Ran He, Zhenan Sun, Tieniu Tan   \\
    {$^1$}School of Artificial Intelligence, University of Chinese Academy of Sciences, Beijing, China \\
    {$^2$}Center for Research on Intelligent Perception and Computing, CASIA, Beijing, China\\
    {$^3$}National Laboratory of Pattern Recognition, CASIA, Beijing, China\\
    {$^4$}Center for Excellence in Brain Science and Intelligence Technology, CAS, Beijing, China\\
    \texttt{ huaibo.huang@cripac.ia.ac.cn} \\
    \texttt{  \{lingxiao.song, rhe, znsun, tnt\}@nlpr.ia.ac.cn }
}

\begin{document}

\maketitle

\begin{abstract}
% hr's version
A capsule is a group of neurons whose activity vector models different properties of the same entity. This paper extends the capsule to a generative version, named variational capsules (VCs). Each VC produces a latent variable for a specific entity, making it possible to integrate image analysis and image synthesis into a unified framework. Variational capsules model an image as a composition of entities in a probabilistic model. Different capsules’ divergence with a specific prior distribution represents the presence of different entities, which can be applied in image analysis tasks such as classification. In addition, variational capsules encode multiple entities in a semantically-disentangling way. Diverse instantiations of capsules are related to various properties of the same entity, making it easy to generate diverse samples with fine-grained semantic attributes. Extensive experiments demonstrate that deep networks designed with variational capsules can not only achieve promising performance on image analysis tasks (including image classification and attribute prediction) but can also improve the diversity and controllability of image synthesis.

\end{abstract}

\section{Introduction}

%第一段：图像分析（目标识别，属性预测等）和合成（条件GAN,VAE等）两个任务的相关性，各自的难点。
%第二段：capsule, VAE, Condtional Generation
%第三段：方法简介
%第四段：Contribution
%As the famous physicist Richard Feynman said, "What I cannot create, I do not understand".

With recent advances in deep learning, tremendous success has been achieved in variety of application domains, including image analysis and synthesis. Image analysis usually refers to extracting information from an image~\cite{zhang2017image} using discriminative models, %such as image classification-related tasks
 while image synthesis aims to produce image samples following an assigned distribution via generative models. These two tasks are highly interconnected and are expected to complement and promote each other.
%Take the popular face synthesis as an instance, powerful face recognition models are vital in guiding generating identity-preserving face images.
Numerous methods attempt to utilize both analysis blocks (e.g., classifiers) and synthesis blocks (e.g., autoregressive models~\cite{van2016pixel,van2016conditional}, VAEs~\cite{kingma2013auto,rezende2014stochastic} and GANs~\cite{goodfellow2014generative}).
In these approaches, analysis blocks are employed to produce controllable conditions for synthesis blocks~\cite{perarnau2016invertible,wang2018high}, or serve as constraints to regularize the target properties of generated samples~\cite{Ledig_2017_CVPR}.
 Nevertheless, in most circumstances the analysis and synthesis blocks are trained in a disjointed way, which may be not an optimal solution for tackling these two tasks simultaneously. %, and few of existing methods integrate the analysis and synthesis tasks into a unified framework.
 It is still a challenge to build a unified framework for image analysis and synthesis, in which these two tasks can collaborate and assist each other. %integrating image analysis and synthesis tasks into a unified framework.%

To alleviate this challenge, we present a new method, called variational capsules (VCs), to model images in a unified discriminative and generative framework. Capsules, which were originally introduced by Hinton et al.~\cite{hinton2011transforming,sabour2017dynamic}, are groups of neurons whose activity vector represents various properties of a particular entity. The proposed variational capsules are a new version of capsules, which use the divergence of each capsule with a prior distribution rather than the length of the activity vector to represent the probability that an entity exists. % and the capsule's specific instantiation to represent the properties of the entity.
Variational capsules model an image as a composition of entities in a probabilistic model, which maps the existing entities into the posterior that matches the prior approximately. Compared with the capsules in ~\cite{sabour2017dynamic}, variational capsules can be drawn from the prior distribution, which extends them from a purely discriminative model to a joint discriminative-generative model.

%In order to take full advantages of Variational Capsules, we propose a VAE-like networks in an auto-encoding architecture.
As illustrated in Figure~\ref{fig:vcs}, our framework takes a VAE-like architecture, and it comprises two parts: an encoder mapping the input images into variational capsules and a generator (or decoder) generating images from masked variational capsules. In the training phase, the encoder aims to detect or classify the existing entity and make the active capsules match the prior distribution, while the decoder tries to reconstruct the input image from the active vectors. In the testing phase, the encoder can be used to analyze the input images via the predicted capsules, and the decoder can synthesize a new sample by drawing samples from the prior distribution. In addition, when dealing with multiple entities, taking attributes as an example, the model can map attribute-aware information into disentangling semantic representations, which makes it possible to synthesize or edit images in a more controllable way with larger diversity. As shown in Figure~\ref{fig:celeba_swap}, the presented model can generate various styles of glasses while preserving other attributes in face synthesis or editing, while most of the other models~\cite{yan2016attribute2image,perarnau2016invertible,lample2017fader} can only generate one style of glasses in this case.

Our contribution is four-fold. i) We present a new version of capsules, variational capsules, that model images as a composition of entities in a probabilistic model. ii) We present a unified framework of image analysis and synthesis, in which image synthesis is helpful for improving the prediction accuracy of image analysis, meanwhile image analysis provides semantic representations for image synthesis. iii) The proposed method provides a new technique for conditional image generation by mapping an image to disentangling semantic representations, which improves the interpretability and diversity of image synthesis. iv) The experiments demonstrate that the proposed method achieves promising or even state-of-the-art performance in some typical image analysis tasks, such as classification and attribute prediction, and outperforms state-of-the-art methods in the diversity and controllability of image synthesis.

\section{Background}

As our work is mostly related to capsules, variational autoencoders (VAEs) and conditional image generation, we start with a brief review of them.

%\subsection{Capsules}

\textbf{Capsules} Hinton et al.~\cite{hinton2011transforming} introduce capsules to represent properties of an image and propose transforming auto-encoder to learn and manipulate an image with capsules. Sabour et al.~\cite{sabour2017dynamic} use the length of a capsule's activity vector to represent the probability of an entity and design an iterative routing-by-agreement mechanism to improve the performance of capsule networks. Hinton et al.~\cite{hinton2018matrix} propose a matrix version of capsules with EM routing. Our work can be seen as a new version of capsules that uses a different metric to represent the presence of an entity. It extends capsules to generative models that are capable of producing new samples.

%It is noted that the proposed variational capsules is trained with convolutional neural networks without routing, as 这句话在conclusion里提

\textbf{Variational Autoencoder (VAE)}~\cite{kingma2013auto,rezende2014stochastic} is one of the most promising generative models for its theory elegancy, stable training and nice manifold representations. VAE consists of two models: a generative model \( p_\theta (x|z) \) to synthesize the visible data \(x\) from the latent code \(z\) and an inference model \( q_\phi (z|x)\) to map the visible data \(x\) to the latent \(z\) which matches to a prior \(p(z)\). The object of VAE is to maximize the variational lower bound (or evidence lower bound, ELBO) of \(p_\theta (x)\):
\begin{equation}\label{eq_vae}
  log p_\theta (x) \geq E_{q_\phi (z|x)} \log p_\theta (x|z) - D_{KL} (q_\phi (z|x) || p(z)).
\end{equation}
The first term in the ELBO aims to reconstruct the input data \(x\) from the posterior \(q_\phi (z|x)\) and the second term aims to make the posterior \(q_\phi (z|x)\) match the prior \(p(z)\). Following the original VAEs~\cite{kingma2013auto}, let the prior \(p(z)\) be the centred isotropic multivariate Gaussian \(N(0,I)\) and the posterior \(q_\phi(z|x) = N(z; \mu, \sigma^2)\), then the KL-divergence term, given \(N\) data samples, can be computed as:
\begin{equation}\label{eq_kl}
  D_{KL}(q_\phi (z|x) || p(z)) =  \frac{1}{2} \sum_{i=1}^{N} (1+ \log (\sigma_i^2) - \mu_i^2 - \sigma_i^2).
\end{equation}
%There are two terms in the ELBO: the first one aims to reconstruct the input data \(x\) from the posterior \(q_\phi (z|x)\) and the second aims to make the posterior \(q_\phi (z|x)\) match to the prior \(p(z)\).

%\textbf{Image classification}

\textbf{Conditional image synthesis} There are mainly two forms of conditional image synthesis according to the provided condition. One is to generate new images \(p_\theta(x|z,c)\) from a prior \(p(z)\) and given conditions \(c\) such as object category, attribute, caption, etc. This task is often implemented by typical generative models with the combination inputs of the latent code \(z\) and the condition \(c\), including CGANs~\cite{mirza2014conditional}, CVAEs~\cite{yan2016attribute2image,Bao_2017_ICCV}, conditional PixelCNN~\cite{van2016conditional}, etc. The other one is to generate new versions \(p_\theta(y|x,c)\) of an input image \(x\) according to the given conditions \(c\), which is also called image transformation (manipulation or editing)~\cite{lample2017fader,perarnau2016invertible,shu2017neural,bao2018towards}.

In conditional image synthesis, the conditions are mostly given or learned as binary codes~\cite{mirza2014conditional,yan2016attribute2image,Bao_2017_ICCV,lample2017fader,perarnau2016invertible} (to indicate category, attribute, caption, etc) or embedding features~\cite{shu2017neural,bao2018towards}. Lample et al.~\cite{lample2017fader} learn attribute-invariant features through adversarial learning and modify an image by sliding the values of the binary attributes. Bao et al.~\cite{bao2018towards} disentangle the identity and attribute features from a face image and map the attribute information into the prior \(N(0,I)\). Compared with the existing methods, our method can learn semantically-disentangling embeddings at a fine-grained level. Taking attribute-guided image synthesis as an example, our method can map the input image into disentangling attribute-aware embeddings, i.e., each attribute is embedded into different capsules, which makes it possible to generate various styles for each attribute while preserving other characteristics of the image.

%1）Capsule
%2) VAE
%3)Conditional Generation

\section{Approach}

\begin{figure}[t]
\centering
\includegraphics[width=0.98\linewidth]{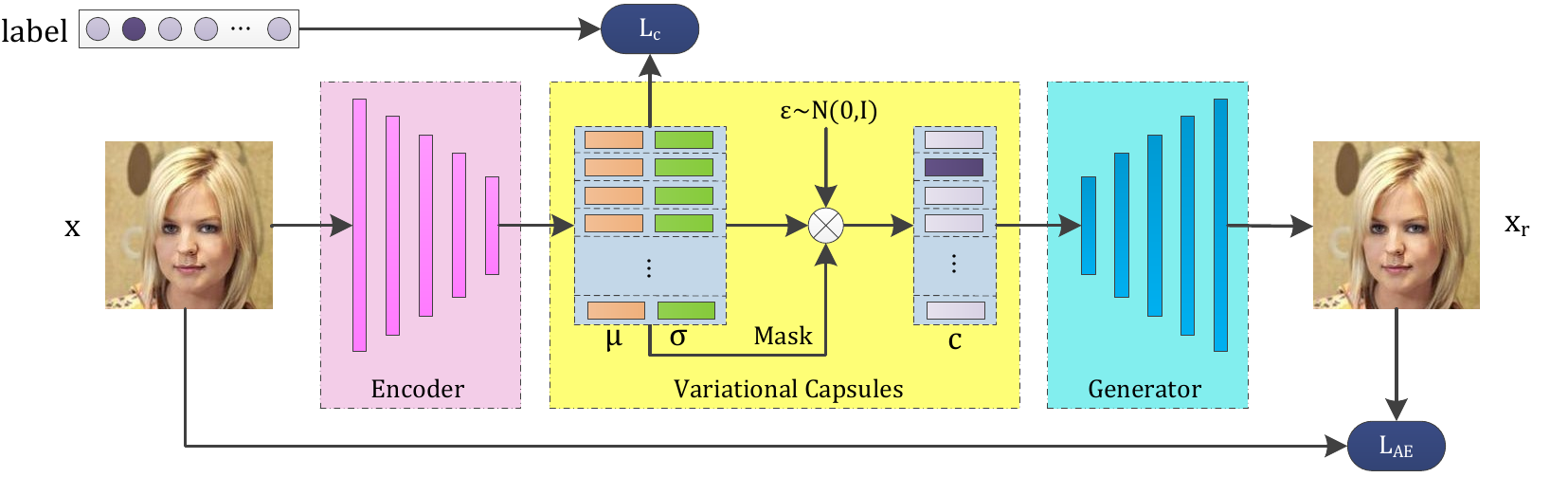}
  \caption{Illustration of the VCNs. The network consists of two parts: an inference model (encoder) to map an input image \(x\) into the posterior \(N(\mu, diag(\sigma^2))\) and a generator model to produce the output image \(x_r\) from the masked capsules. The posterior is trained to match the prior $p(z)$ for the active capsules, and to deviate from the prior \(p(z)\) for the nonactive capsules. The input capsules of the generator are sampled from the posterior (or from the prior when sampling new images) using the reparameterization trick with a mask to indicate the present entities. }
  \label{fig:vcs}
\end{figure}

To analyze and synthesize images in a unified framework, the proposed VCs are expected to have two properties: one is that the active vector of a capsule is able to indicate the existing probability of an entity in an image; the other is that the capsule follows a known prior distribution to allow sampling of new capsules from the prior distribution to generate new images. In this section, we first describe how to design such variational capsules, followed by the training details of the VC networks (VCNs) and their applications in image synthesis.

\subsection{Variational Capsules}

The capsules proposed in~\cite{sabour2017dynamic} use the length of the instantiation vector to represent the probability of the existing entity. To facilitate the sampling of new capsules, we design variational capsules in a probabilistic manner that the active capsules follow a known prior distribution while the nonactive ones do the opposite. Following VAEs~\cite{kingma2013auto}, we select the KL-divergence as the metric to indicate the degree how two distributions match to each other. Hence, the KL-divergence of each capsule with the prior distribution represents the probability that a capsule's entity exists, i.e., the capsule corresponding to the existing entity has a small KL-divergence with the prior while those corresponding to the non-existing entities have large KL-divergences with the prior distribution.

Following the original VAEs~\cite{kingma2013auto}, the prior \(p(z)\) is assumed to follow isotropic multivariate Gaussian distribution, i.e., \(p(z) \sim N(0, I)\), while the proposed capsule \(q_\phi(c|x)\) follows multivariate Gaussian distribution whose mean and covariance are parameterized by \(N(\mu(x), diag(\sigma^2(x)))\). The KL-divergence of each capsule \(c\)  with the prior \(p(z)\), i.e., \(D_{KL}(q_\phi(c|x)||p(z))\), can be computed using Eq.~(\ref{eq_kl}). Let \(L_{KL}(c)\) denote the above divergence, we use a separate margin loss \(L_k\) for each capsule \(c_k\) (where \(k\) indicates the index of the capsule), which is defined as:
\begin{equation}\label{e3}
  L_k = T_k~ L_{KL}(c_k) + \lambda~ (1 - T_k)~  [m - L_{KL}(c_k)]^+,
\end{equation}
where \(T_k=1\) if and only if the entity (such as a category or an attribute) exists;  \([\cdot] = max(0,\cdot)\), \(m\) is a positive margin; and \(\lambda\) is a down-weighting coefficient. The total loss \(L_c\) is the sum of the losses of all the capsules, i.e., \(L_c = \sum_{k} L_k\).

\subsection{Training VCNs}

As illustrated in Figure~\ref{fig:vcs}, VCNs are designed like VAEs~\cite{kingma2013auto} with an auto-encoding architecture. The VCNs contain two modules: an inference model (Encoder) to map an input image \(x\) into an approximate posterior matching the prior distribution, and a generator model (Generator) to generate samples from the capsules. During training, the proposed variational capsules \(C\) in Figure~\ref{fig:vcs} are sampled from the posterior \(N(\mu(x), diag(\sigma^2(x)))\) where \(\mu(x)\) and \(\sigma(x)\) are the output vectors of the encoder. Following VAEs~\cite{kingma2013auto}, the capsules are sampled using the reparameterization trick, i.e., \(c = \mu + \varepsilon \odot \sigma\), where \(\varepsilon \sim N(0, I)\) is a random vector and \(\odot\) means the element-wise multiplication.
Besides, we take masked capsules as the input of the generator; i.e., all the capsules are set to zero, %except the active capsule whose KL-divergence with the prior distribution is minimal among the capsules.
except for the active capsule that has the minimal KL-divergence with the prior distribution.

Similar to the capsules in~\cite{sabour2017dynamic}, we use an additional reconstruction loss \(L_{AE}\) to encourage the proposed capsules to capture the entity's instantiation details of the input image. The auto-encoding loss \(L_{AE}\) is a classic pixel-wise mean squared error (MSE) which estimates the element-wise reconstruction quality given \(N\) data samples:
\begin{equation}\label{eq_mse}
  L_{mse} =  \frac{1}{2} \sum_{i=1}^{N} \| x_{r,i} - x_i \|^{2}_{F},
\end{equation}
where \(x_{r,i}\) is the reconstruction of the \(i\)-th data \(x_i\). It is noted that the exact choice of the auto-encoding loss is not fundamental for the proposed methods. For example, cross entropy loss may be more suitable for binary images such as digits in MNIST~\cite{lecun1998gradient}.

In addition, when dealing with high-resolution images such as faces in CelebA~\citep{Liu_2015_ICCV}, additional losses could be adopted to boost synthesis performance~\cite{Ledig_2017_CVPR,Bao_2017_ICCV,larsen2016autoencoding}, such as the adversarial loss in GANs~\cite{goodfellow2014generative} to promote generating sharp images, and the perceptual loss~\cite{bruna2016super,johnson2016perceptual} to regularize high-level semantic property. % capture more details.
When the adversarial loss is employed, an extra discrimination model (Discriminator) is introduced to compete with the above mentioned encoder and generator as in traditional GANs. Concretely, we take the form of the adversarial loss $L_{adversarial}$ as in LSGAN~\cite{mao2017least} to get better convergence and higher image quality.
\begin{equation}\label{eq_adv}
  \mathop {\min }\limits_G \mathop {\max }\limits_D {L_{adversarial}} = {E_{p\left( x \right)}}{\left[ {D(x)} \right]^2} + {E_{p\left( z \right)}}{\left[ {1 - D\left( {G(z)} \right)} \right]^2}.
\end{equation}

The final objective takes the following form:
\begin{equation}\label{eq_total-loss}
  L_{total} = L_{c} + \alpha~ L_{mse} + \beta~ L_{adversarial}
\end{equation}
where \(\alpha\), \(\beta\) are weighting coefficients to balance the importance of the losses.

\subsection{Image Synthesis with VCs}

%Variational capsules encode an input image into disentangling semantic representations that each capsule corresponds with a specific entity, which makes it easy to control the properties of the synthesised image, such as object category or attributes. Besides, as the capsules follow a known prior, different sampling of an individual capsule would cause various instantiations of the corresponding entity.

There are three steps to generate a new sample with variational capsules: (1) determine the mask \(M\) according to the expected entity, such as the object category or attribute of the generated image; (2) sample the variational capsules \(C\) from the prior distribution \(p(z)\), or from the posterior \(q_\phi(z|x)\) given an input image \(x\); (3) produce the output image from the masked capsules \(M \odot C\) using the generator. Through these steps, it is possible to control the synthesized images in a fine-grained way. We can modify the mask to change the expected entities in the output image and can use different instantiations of an activity capsule to generate large variations of a specific entity.

The advantages of the proposed method is mostly remarkable when dealing with the images that contain multiple entities. Variational capsules encode a single image into disentangling semantic representations in which each capsule corresponds with a specific entity. With the semantically-disentangling representation, it is easy to control the properties of the synthesised image. Taking attribute-guided image generation as an example, an input image is modeled as a composition of multiple attribute-related capsules. We can synthesize a completely new image with the latent representation sampled from the prior distribution. We can also modify a given image or a synthesized image in an attribute-level approach so that it is easy to generate various styles of a single attribute while preserving the properties of other attributes. For example, the proposed VCNs can synthesize images with various styles of bangs or glasses (see in Figure~\ref{fig:celeba_swap}). In contrast, most other works can only change the existence of an attribute~\cite{yan2016attribute2image,perarnau2016invertible} or the intensity of the attribute~\cite{lample2017fader,Bao_2017_ICCV,bruna2016super}, but can hardly provide different styles of the attribute.

%1)综述
%2）公式
%3）网络及损失函数

\section{Experiments}

We implement experiments on two datasets: attribute prediction and synthesis on CelebA~\cite{Liu_2015_ICCV} , digit classification and generation on MNIST~\cite{lecun1998gradient}, to evaluate the performance of the presented method on image analysis and synthesis.

%\subsection{Implementation Details}
%Image sizes for CelebA and MNIST are $256\times256$ and $28\times28$ respectively.

%Our encoder and decoder network

%We adapt our architectures from Johnson et al. [19].
%We use 6 blocks for 128 × 128 training images, and 9 blocks for 256 × 256 or higher-resolution training images. Below, we follow the naming convention used in the Johnson el al.’s Github repository.

%Reflection padding was used to reduce artifacts. Rk denotes a residual block that contains two 3 × 3 convolutional layers with the same number of filters on both layer. uk denotes a 3 × 3 fractional-strided-Convolution-BatchNorm-ReLU layer with k filters, and stride 1 2.
%\subsubsection{Training details}

%\texttt{c5s1-16d}, \texttt{D}, \texttt{R32}, \texttt{D}, \texttt{R64}, \texttt{D}, \texttt{R128}, \texttt{D}, \texttt{R256}, \texttt{D}, \texttt{R512}, \texttt{D}, \texttt{c1s1-10240}, \texttt{c4s1-5120}

\subsection{Experiments on CelebA}
\textbf{The CelebA database}~\cite{Liu_2015_ICCV} consists of 202,599 celebrity images with large variations in facial attributes. These images are obtained from
unconstrained environments and annotated with 40 attributes. The standard split for CelebA is employed in our experiments, where 162,770 images for training, 19,867 for validation and 19,962 for testing. Following the image pre-processing method in~\cite{lample2017fader}, we use the aligned version of CelebA in our experiments. Images are firstly center cropped to $178\times 178$ and then resized to $256 \times 256$ before fed in our networks.

%\textbf{Network architecture.}
We treat the attribute prediction as a multi-task binary classification problem. For each attribute, we train a classifier with two outputs that model the active/nonactive status of this attribute. Following the original VAEs, each output is formed by a pair of variational capsules that represent the mean and covariance of the posterior distribution, respectively. In our attribute prediction experiment, the dimension of variational capsule is set to 32. Therefore, in total the encoder has $40\times2\times2$ variational capsule outputs, and each capsule is a 32-D vector.
The decoder receives $40\times2$ capsules as input, which are sampled from the posteriors, i.e., the outputs of the encoder.

Let $\texttt{C5s1-k}$ denote a $5\times5$ \text{Convolution-BatchNorm-ReLU} layer with \texttt{k} filters and stride 1. $\texttt{d}$ denotes an average pooling layer with $2\times2$ kernel size and stride 2. $\texttt{u}$ denotes an upsampling layer with sale factor 2. $\texttt{Rk}$ denotes a residual block that contains two $3 \times 3$ \text{Convolution-BatchNorm-ReLU} layers with \texttt{k} filters, and an extra $1 \times1$ convolution layer with \texttt{k} filters in the identity path when the input channel does not equal \texttt{k}. $\texttt{Fk}$ denotes a fully-connected layer with output dimension \texttt{k}. If one convolutional layer is followed by denotation \texttt{gk} , the convolutional filters in this layer are separated into \texttt{k} groups.
The encoder architecture is:
\texttt{C5s1-16d}, \texttt{R32d}, \texttt{R64d}, \texttt{R128d}, \texttt{R256d}, \texttt{R512d}, \texttt{C1s1-10240}, \texttt{C4s1g40-5120}.
The decoder architecture is:
\texttt{F8192}, \texttt{R512u}, \texttt{R256u}, \texttt{R128u}, \texttt{R64u}, \texttt{R32u}, \texttt{R16u}, \texttt{R16}, \texttt{c5s1-3}.

An extra multi-scale discriminator is employed to differentiate between natural and synthesized samples, with which adversarial training is conducted. All these sub-networks are trained jointly with a batch size of 64 and a learning rate of $2\times10^{-4}$. In our experiment, we empirically set the trade-off parameters for reconstruction loss and adversarial loss to 0.025 and 10, respectively.

\subsubsection{Attribute Prediction}

\begin{table*}
  \centering\tiny
  \caption{Attribute prediction accuracies on CelebA. Attributes are numbered from 1 to 40 in alphabetical order.}
  \label{tab:celeba}
  \setlength{\tabcolsep}{0.55mm}{
  \begin{tabular}{c|ccccccccccccccccccccc}
     \hline\noalign{\smallskip}
   Approach&1&2&3&4&5&6&7&8&9&10&11&12&13&14&15&16&17&18&19&20&21 \\
   % Approach&\rotatebox{90}{5 Shadow}&\rotatebox{90}{Arch. Eyebrows}&\rotatebox{90}{Attractive}&\rotatebox{90}{Bags Un. Eyes}&\rotatebox{90}{Bald}&\rotatebox{90}{Bangs}&\rotatebox{90}{Big Lips}&\rotatebox{90}{Big Nose}&\rotatebox{90}{Black Hair}&\rotatebox{90}{Blond Hair}&\rotatebox{90}{Blurry}&\rotatebox{90}{Brown Hair}&\rotatebox{90}{Bushy Eyebrows}&\rotatebox{90}{Chubby}&\rotatebox{90}{Double Chin}&\rotatebox{90}{Eyeglasses}&\rotatebox{90}{Goatee}&\rotatebox{90}{Gray Hair}&\rotatebox{90}{Heavy Makeup}&\rotatebox{90}{High Cheekbones}&\rotatebox{90}{Male} \\

     \hline\noalign{\smallskip}
     LNets+ANet~\cite{Liu_2015_ICCV} & 91.00 & 79.00 & 81.00 & 79.00 & 98.00 & 95.00 & 68.00 & 78.00 & 88.00 & 95.00 & 84.00 & 80.00 & 90.00 & 91.00 & 92.00 & 99.00 & 95.00 & 97.00 & 90.00 & 87.00 & 98.00 \\
     MOON~\cite{rudd2016moon} & 94.03& 82.26& 81.67& 84.92& 98.77& 95.80 &71.48& 84.00& 89.40& 95.86& 95.67& 89.38 &92.62& 95.44& 96.32 &99.47& 97.04& 98.10 & 90.99 & 87.01& 98.10 \\
     MCNN+AUX~\cite{hand2017attributes} & 94.51& 83.42 &83.06& 84.92& 98.90& 96.05& 71.47& 84.53& 89.78& 96.01& 96.17& 89.15 &92.84 &95.67& 96.32& 99.63 & 97.24 &98.20& 91.55& 87.58& 98.17\\
     PaW~\cite{ding2017deep} & 94.64& 83.01 &82.86 &84.58& 98.93& 95.93& 71.46 &83.63& 89.84 &95.85& 96.11 &88.50& 92.62 &95.46& 96.26& 99.59 &97.38& 98.21& 91.53 &87.44 &\textbf{98.39} \\
     %Ours1 & 94.92 & 84.08 & 82.76 & 85.59 & 99.09 & 96.04 & 71.70 & 84.27 & 89.95 & 96.08 & 96.24 & 89.62 & 92.99 & 95.79 & 96.56 & 99.62 & 97.65 & 98.30 & 91.75 & 87.90 & 98.30 \\
     %\hline
     Ours( w/o recon.) & 94.64 & 84.10 &83.01& 85.05 &98.90 & 95.98 &71.43 & \textbf{84.99} &89.58 & 96.06  & \textbf{96.26}  & 89.06  & 93.01  & \textbf{95.96}  & \textbf{96.60}  & 99.69  & 97.52  & 98.28  & 91.57  & 87.71 &98.32\\
     Ours  &\textbf{94.88} & \textbf{84.15} & \textbf{83.19} & \textbf{85.69}& \textbf{99.05}& \textbf{96.09} & \textbf{71.75}& 84.95 & \textbf{90.23}& \textbf{96.28}& \textbf{96.26} & \textbf{90.00} & \textbf{93.06} &  95.66 &  96.58 &  \textbf{99.70} &  \textbf{97.66} &  \textbf{98.37} &  \textbf{92.06} &  \textbf{87.80} &  98.37 \\
    \hline\noalign{\smallskip}

 Approach&22&23&24&25&26&27&28&29&30&31&32&33&34&35&36&37&38&39&40&&Avg \\
%Approach&\rotatebox{90}{Mouth S. O.}&\rotatebox{90}{Mustache}&\rotatebox{90}{Narrow Eyes}&\rotatebox{90}{No Beard}&\rotatebox{90}{Oval Face}&\rotatebox{90}{Pale Skin}&\rotatebox{90}{Pointy Nose}&\rotatebox{90}{Reced. Hairline}&\rotatebox{90}{Rosy Cheeks}&\rotatebox{90}{Sideburns}&\rotatebox{90}{Smiling}&\rotatebox{90}{Straight Hair}&\rotatebox{90}{Wavy Hair}&\rotatebox{90}{Wear. Earrings}&\rotatebox{90}{Wear. Hat}&\rotatebox{90}{Wear. Lipstick}&\rotatebox{90}{Wear. Necklace}&\rotatebox{90}{Wear. Necktie}&\rotatebox{90}{Young}&&Avg \\
    \hline\noalign{\smallskip}
     LNets+ANet~\cite{Liu_2015_ICCV} & 92.00 & 95.00 & 81.00 & 95.00 & 66.00 & 91.00 & 72.00 & 89.00 & 90.00 & 96.00 & 92.00 & 73.00 & 80.00 & 82.00 & 99.00 & 93.00 & 71.00 & 93.00 & 87.00 &  & 87.30 \\
     MOON~\cite{rudd2016moon}& 93.54& 96.82& 86.52& 95.58& 75.73 &97.00 &76.46 &93.56& 94.82& 97.59& 92.60& 82.26& 82.47& 89.60& 98.95& 93.93& 87.04 & 96.63 & 88.08 && 90.94\\
     MCNN+AUX~\cite{hand2017attributes} & 93.74 &96.88 &87.23& 96.05& 75.84& 97.05& 77.47& 93.81& 95.16& 97.85& 92.73 &83.58& 83.91 &90.43& 99.05& 94.11 & 86.63 &96.51 &88.48 & & 91.29 \\
     PaW~\cite{ding2017deep} & \textbf{94.05} & 96.90 & 87.56 & 96.22 & 75.03 & \textbf{97.08 }& 77.35 & 93.44 & 95.07 & 97.64 & 92.73 & 83.52 & 84.07 & 89.93 & 99.02 & 94.24 & 87.70 & 96.85 & 88.59&  & 91.23 \\
     Ours( w/o recon.) & 94.02 & 96.95 & 87.53 & \textbf{96.57}& \textbf{75.96} & 97.06  & \textbf{77.70} & 93.85 & 94.94 & 97.84 & 93.16 & 83.69 & 84.30 & 90.81 & 99.07  & \textbf{94.37}  & 86.21  & 96.42  & \textbf{88.72}  &    & 91.42  \\
     %Ours1 & 93.98 & 96.99 & 87.40 & 96.23 & 74.84 & 97.13 & 77.19 & 93.90 & 95.37 & 97.95 & 93.04 & 83.06 & 83.08 & 90.54 & 99.04 & 94.31 & 86.60 & 96.26 & 88.20&& \textbf{91.36}  \\
     Ours  & 93.98 & \textbf{96.97} &\textbf{87.70} & 96.43 &73.82& \textbf{97.08} & 76.50 & \textbf{93.99}& \textbf{95.24} & \textbf{97.97} & \textbf{93.17} & \textbf{84.39} & \textbf{84.40} & \textbf{90.84} & \textbf{99.13} & 94.14 & \textbf{87.81} &  \textbf{97.15} &  88.65 & &  \textbf{91.53}\\

     \hline
   \end{tabular} }
   \vspace{-0.02\linewidth}
\end{table*}

To demonstrate the capacity of our method in image analysis, we conduct attribute prediction in CelebA. The classification accuracies are reported in Table~\ref{tab:celeba}. Our method obtains an average accuracy of $91.36\%$, outperforming the baseline method LNets+ANet~\cite{Liu_2015_ICCV} by over $4\%$. Besides, VCNs perform better than PaW~\cite{ding2017deep} which uses multiple networks, and MCNN+AUX~\cite{hand2017attributes} which elaborately categorizes the attributes into different groups. Adding the reconstruction boosts the prediction accuracy, suggesting that image synthesis is helpful for learning discriminative representation. %Particularly, we observe that our method performs better in predicting attributes that explicitly affect the visual appearance of face images, such as `$\text{5\_o\_Clock\ Shadow}$', `Bangs' and `Goatees'. As for the attributes which concern implicit cues, e.g. `Attractive' and `Male', the accuracies of our method are not that high, but still satisfactory comparing with other approaches.
Particularly, we observe that the reconstruction loss contributes a lot in predicting attributes which explicitly affect the visual appearance of face images, such as `Heavy Makeup', `Rocy Cheeks' and `Straignt Hair'. %As for the attributes which concern implicit cues, e.g. `Attractive' and `Male', the accuracies of our method are not that high, but still satisfactory comparing with other approaches.

\subsubsection{Face Synthesis} In this part, we provide experiments for generating face images from latent representations. Attribute-conditioned image generation, facial attribute swapping and attribute interpolating are conducted to demonstrate our models' ability in synthesizing face images with great diversity and fine-grain controllability.

\begin{figure*}[bht]
\begin{center}
\includegraphics[width=0.99\linewidth]{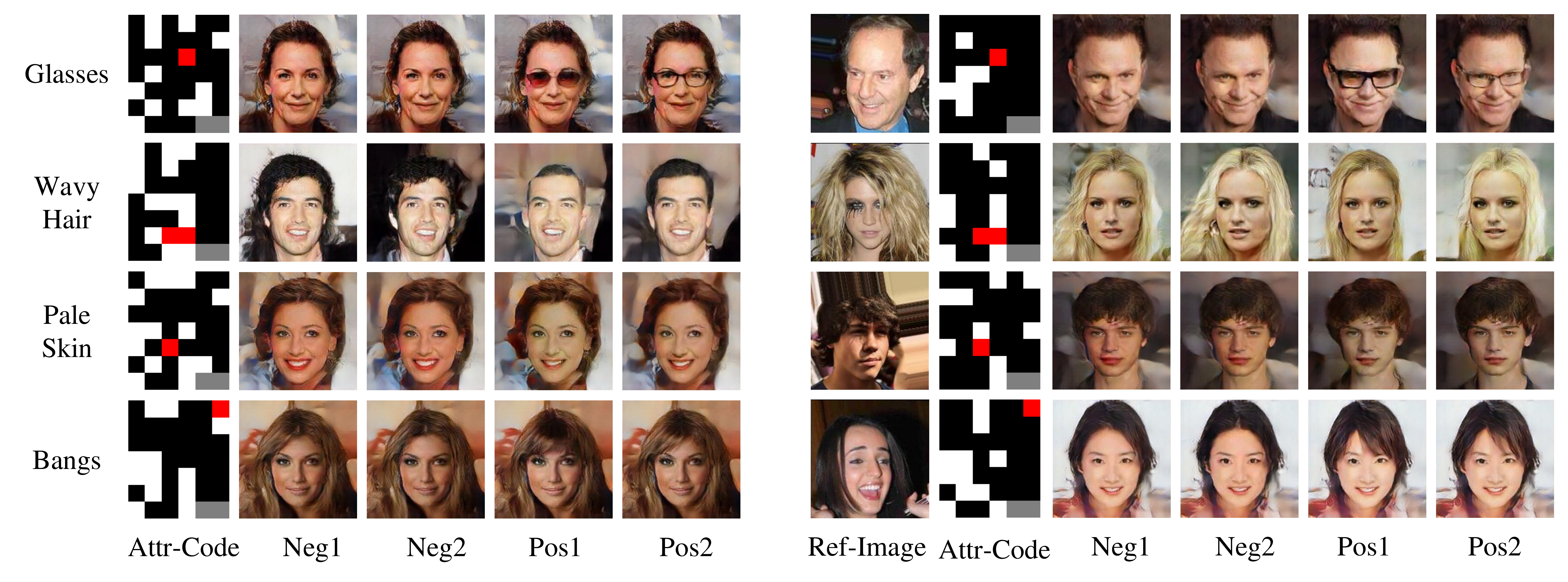}\vspace{-0.04\linewidth}
\end{center}
\caption{
     Attribute-conditioned image generation. Blocks corresponding to the changed attributes are marked with red color. %The attributes are provided respectively by the reference image (Ref-Image) and binary attribute code (Attr-Code, the light and dark blocks represent the negative and positive status of the \(40\) attributes, respectively). We give two variations for each of the negative and positive statuses of each attribute. Best viewed by zooming in the electronic version.
}
\label{fig:celeba_condition}
\vspace{-0.02\linewidth}
\end{figure*}

\textbf{Attribute-conditioned image generation}. Figure~\ref{fig:celeba_condition} shows examples of generating face images from specified attributes.
The left side of Figure~\ref{fig:celeba_condition} shows results of directly synthesizing new faces from latent codes. The latent codes are randomly sampled from prior $N(0,I)$ and then masked accordingly to targeted attributes. In the first column of each example, a binary block image exhibits the exact activation status of the 40 attributes in CelebA, corresponding to the first 40 blocks in $7 \times 6$ matrices. %Latent codes for most variational capsules with only specific capsules resampled.
For each example in the right side, a reference image is involved to generate images sharing the same attributes with it. Concretely, attributes are firstly predicted via the proposed inference model (encoder), then latent codes are sampled according to the prediction results. % from reference images with only specific capsules resampled.
We change the latent code for one attribute in each example by adjusting specific capsules while keeping the rest fixed. Two positive and two negative samples are provided for the changed attribute.
As shown in Figure~\ref{fig:celeba_condition}, all the generated faces are visually plausible and accord with targeted attributes. Attributes of the reference images are well transferred to new generated images, which demonstrates that VCNs perform well in both image analysis and synthesis. Specifically, when we change specific attributes, the rest attributes are well preserved, suggesting our model's ability in learn semantically-disentangling representations.

\begin{figure*}[bht]
\begin{center}
\vspace{-0.01\linewidth}
\includegraphics[width=0.99\linewidth]{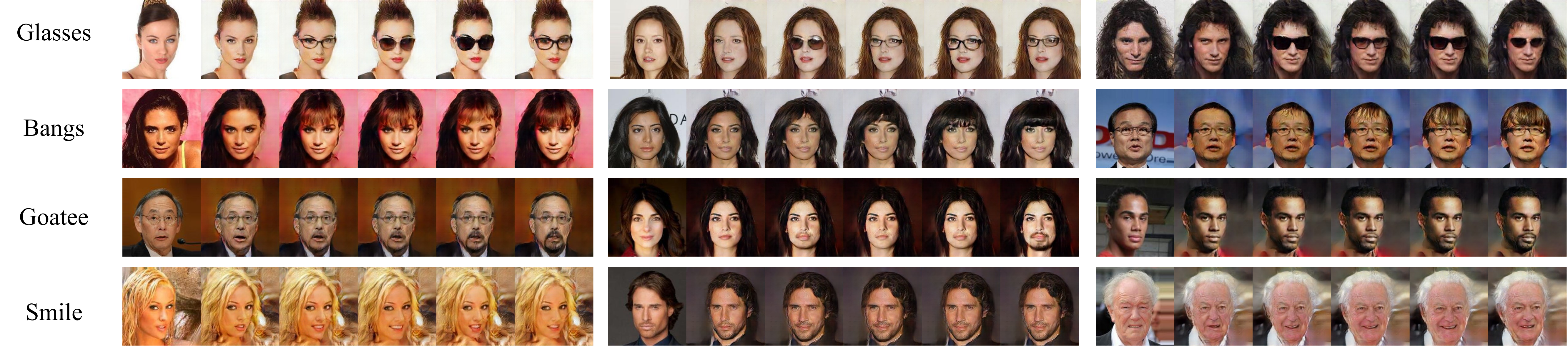}\vspace{-0.03\linewidth}
\end{center}
\caption{
      Swapping the attributes of different faces. From left to right, original faces, reconstruction faces and various attribute-swapping results.%For each identity, the first and second images are the original and reconstruction faces from the CelebA testing set, respectively; the rest four faces are the various versions of the original face by swapping a specific attribute while preserving other attributes. Best viewed by zooming in the electronic version.
}
\vspace{-0.01\linewidth}
\label{fig:celeba_swap}
\end{figure*}

\textbf{Facial attribute swapping.}
Some visual examples of facial attribute swapping are shown in Figure~\ref{fig:celeba_swap}. For each identity, the first and second images are the original and reconstruction faces from the CelebA testing set, respectively. Due to the injection of random noise in the training phase of VCNs, the reconstruction images cannot keep accurate pixel-wise similarity with the original images. But the attributes of reconstruction remain unchanged. The rest four faces are synthesis results by swapping an attribute of the input face while keeping other attributes preserved. These generated images confirm that the proposed VCNs are able to learn semantically disentangling features. In addition, various embodiments of the same attribute can be accessed by resampling the proposed variational capsules. For example, different styles of glasses and bangs are presented in these generated faces.

%\begin{figure*}[bht]
%\begin{center}
%   \subfigure[FaderNet~\cite{lample2017fader}]{\includegraphics[width=0.48\linewidth]{comp_1.pdf}}
%   \subfigure[Ours]{\includegraphics[width=0.48\linewidth]{comp_2.pdf}}\vspace{-0.03\linewidth}
%\end{center}
%\caption{
%     Examples of multi-attribute swap (Gender / Opened eyes / Eye glasses). Left images are the originals. %The first and last columns are the attribute codes for the leftmost and rightmost faces, respectively. The first row is interpolated between the capsules' instantiation parameters with the same attribute label. The second row is interpolated between attribute labels using the same capsules' instantiation parameters. The bottom two rows are interpolated between both the attribute labels and the capsules' instantiation parameters. The learned representations using variational capsules are continuous both for the semantic labels and the specific instantiations.
%}
%\vspace{-0.03\linewidth}
%\label{fig:celeba_inter_multi}
%\end{figure*}

\begin{figure*}[bht]
\begin{center}
\vspace{-0.01\linewidth}
\includegraphics[width=0.99\linewidth]{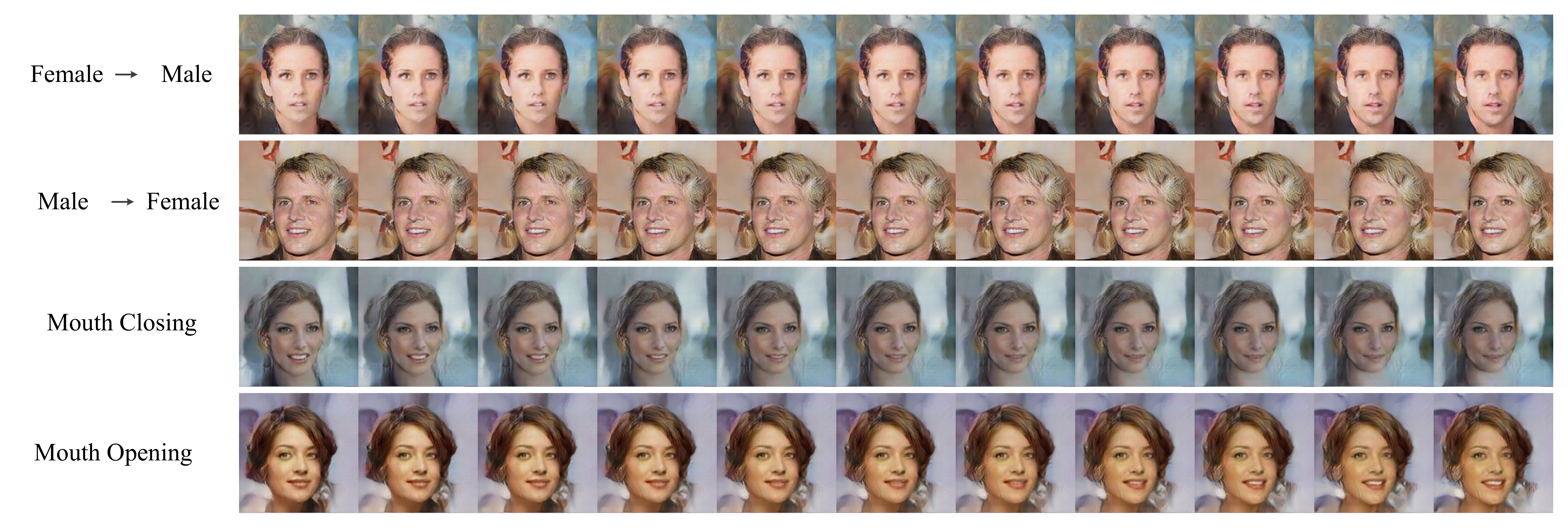}\vspace{-0.03\linewidth}
\end{center}
\caption{
     Interpolation between individual attributes. %The attribute intensity can be modified continuously by the interpolation between the activity and non-activity capsules using a continuous variable. Best viewed by zooming in the electronic version.
}
\vspace{-0.02\linewidth}
\label{fig:celeba_inter_single}
\end{figure*}

\begin{figure*}[bht]
\begin{center}
\vspace{-0.04\linewidth}
\includegraphics[width=0.99\linewidth]{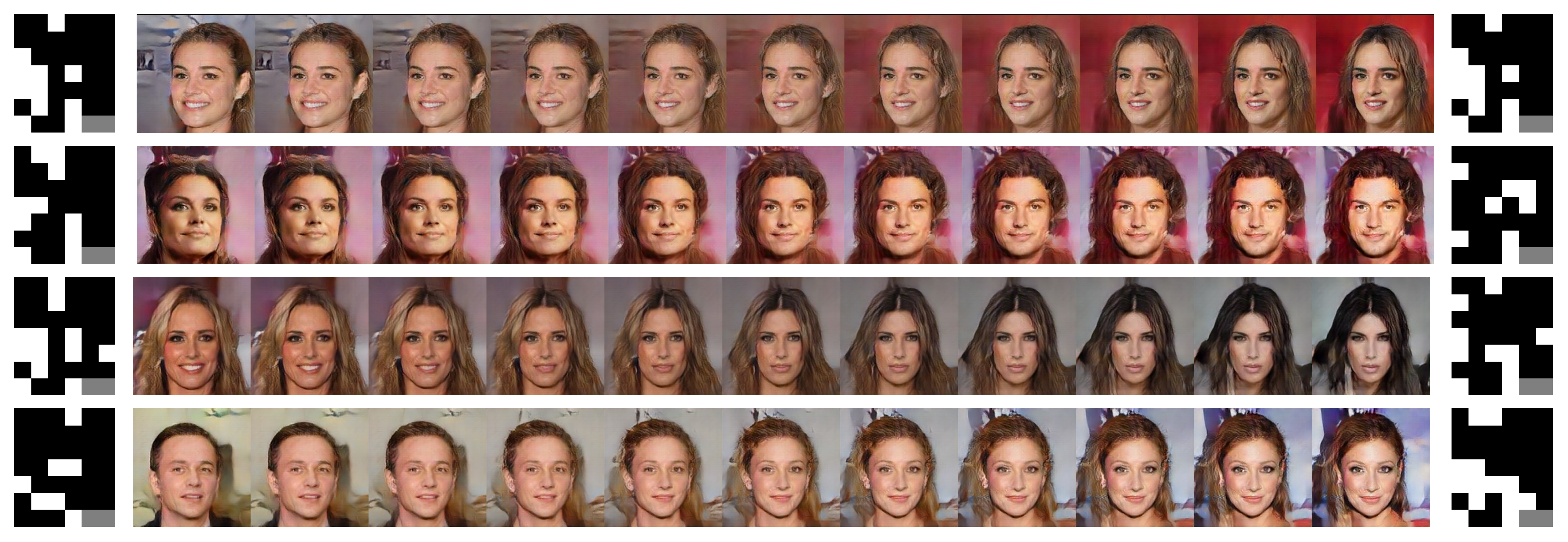}\vspace{-0.03\linewidth}
\end{center}
\caption{
     Interpolation between multiple attributes. %The first and last columns are the attribute codes for the leftmost and rightmost faces, respectively. The first row is interpolated between the capsules' instantiation parameters with the same attribute label. The second row is interpolated between attribute labels using the same capsules' instantiation parameters. The bottom two rows are interpolated between both the attribute labels and the capsules' instantiation parameters. The learned representations using variational capsules are continuous both for the semantic labels and the specific instantiations.
}
\label{fig:celeba_inter_multi}
\vspace{-0.03\linewidth}
\end{figure*}

\textbf{Facial attribute interpolation.}
In this part, we conduct attribute interpolation experiments to show our method's ability in continuously changing facial attributes. Figure~\ref{fig:celeba_inter_single} shows results for single attribute interpolation. Specifically, we change the attribute intensity by linearly interpolating between an activate capsule and a nonactive capsule. Subtle changes exist in contiguous images, while images in both ends differ from each other in greater degree. These results demonstrate the operation-friendliness of our method, as we can easily synthesize facial images with desired attributes and intensities. Apart from single attribute interpolation, results for multi attributes interpolation are also provided (Figure~\ref{fig:celeba_inter_multi}). As mentioned above, diverse samples can be synthesized from the same active capsules with different instantiations. Thus, we interpolate two different instantiations of the same attributes in the first row of Figure~\ref{fig:celeba_inter_multi}, in which faces change gradually meanwhile the attributes are kept unchanged. The rest rows are interpolation results between faces with different attributes. The interpolation results are visual-pleasing, and continuous attributes change can be found in these generated images, confirming our method's capacity in representing facial attributes again.

\subsection{Experiments on MNIST}
\textbf{The MNIST database}~\cite{lecun1998gradient} is a digit dataset with 60,000 training and 10,000 testing images. All images in MNIST dataset are binary images of size $28\times28$, and each image contains a single handwritten digit with the class label from 0 to 9.
%images that have been shifted by up to 2 pixels in each direction with zero padding. No other data augmentation/deformation is used

%\textbf{Network architecture.}
The encoder used in this experiment consists of 3 convolutional layers, followed by a fully-connected layer which produces variational capsules for means and covariances of 10 digit class. The dimension of variational capsules is set to 16 such that the output of encoder is of 320-D.
%We use a similar network architecture as, in which a 3-layer single network is adopted. The final layer of encoder produces 16-D capsules per digit class.% and each of these capsules receives input from all the capsules in the layer below.
%Detailed structures of encoder and decoder are listed as follows.
The encoder architecture is: \texttt{C5s1-256}, \texttt{C5s2-256}, \texttt{C5s2-256}, \texttt{F320}.
The decoder is formed of 3 fully-connected layers, and the detailed network architecture is: \texttt{F512}, \texttt{F1024}, \texttt{F784}.

%MNIST

%分类，重建，生成数字， 生成叠加的数字

%\begin{figure}[t]
%\centering
%\includegraphics[width=0.98\linewidth]{VC.pdf}
%  \caption{Synthesis results of digits. From left to right, input digit images, reconstruction digit images, (x) to (x) generated digit face images, and (x) to (x) interpolated results. }
% \label{fig:mnist}
% \end{figure}

\begin{figure*}[bht]
\begin{center}\vspace{-0.03\linewidth}
   \subfigure[Original]{\includegraphics[width=0.3\linewidth]{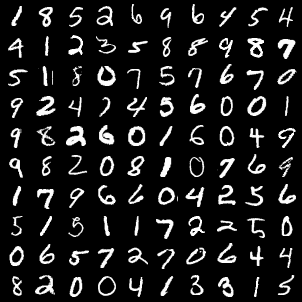}}
   \subfigure[Reconstructions]{\includegraphics[width=0.3\linewidth]{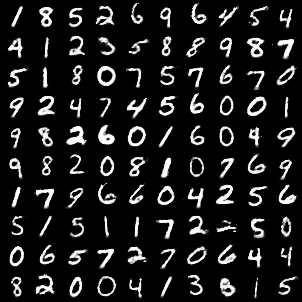}}
   \subfigure[Samples]{\includegraphics[width=0.3\linewidth]{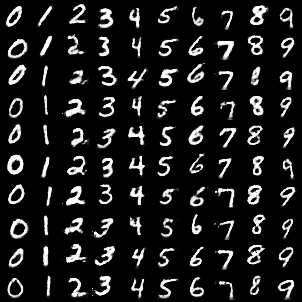}}\vspace{-0.03\linewidth}
\end{center}
\caption{
    Results on the MNIST dataset. From left to right are the original images from the MNIST test set, the reconstruction images and the synthesized new digit samples.
}
\vspace{-0.02\linewidth}
\label{fig:mnist}
\end{figure*}

\begin{figure*}[bht]
\begin{center}\vspace{-0.03\linewidth}
\includegraphics[width=0.9\linewidth]{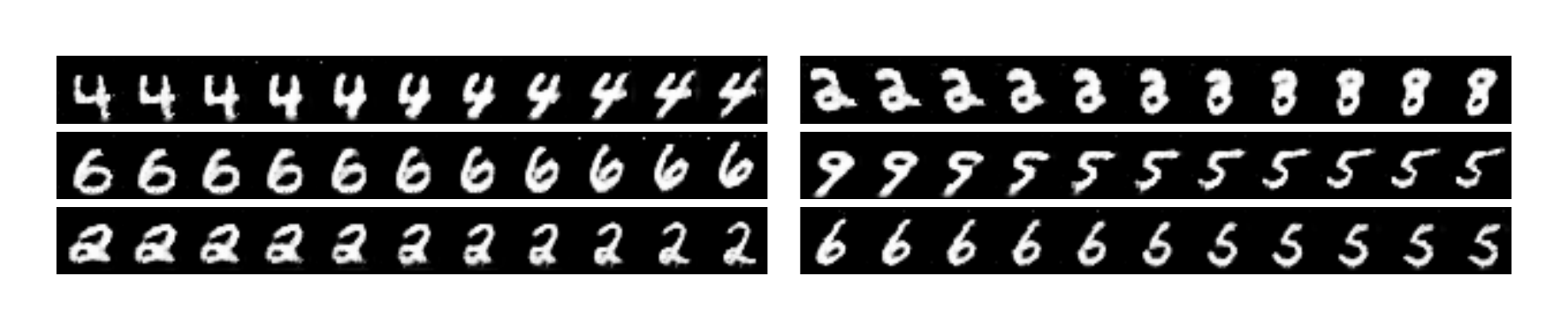}\vspace{-0.05\linewidth}
\end{center}
\caption{
     Interpolation between digit images. %The first and last columns are the attribute codes for the leftmost and rightmost faces, respectively. The first row is interpolated between the capsules' instantiation parameters with the same attribute label. The second row is interpolated between attribute labels using the same capsules' instantiation parameters. The bottom two rows are interpolated between both the attribute labels and the capsules' instantiation parameters. The learned representations using variational capsules are continuous both for the semantic labels and the specific instantiations.
}
\vspace{-0.02\linewidth}
\label{fig:mnist_inter}
\end{figure*}

\textbf{Digit recognition.} We obtain an error rate of $0.36\%$ using the encoder only, and $0.30\%$ when introducing the reconstruction loss. The introducing of auto-encoding architecture helps in improving the classification accuracy.
%For comparison, . % Wan et al.~\cite{wan2013regularization} .
Sabour et al.~\cite{sabour2017dynamic} implement a 3-layer capsule network, and achieve error rates of $0.35\%$ without reconstruction loss and $0.25\%$ with reconstruction loss respectively. Hinton et al.~\cite{hinton2018matrix} get $0.44\%$ error rate with matrix capsules. Dynamic routing mechanism is employed in these capsule networks~\cite{sabour2017dynamic,hinton2018matrix} and plays important roles. However, because of the different way of representing the probability that an entity exists, these routing algorithms in~\cite{sabour2017dynamic,hinton2018matrix} cannot be directly applied in the proposed variational capsules. It will be a potential research direction for our method to explore the design of routing algorithm.

\textbf{Digit synthesis.}
As illustrated in Figure~\ref{fig:mnist}, the reconstructions of variational capsules are robust while keeping only important details.
  Apart from reconstructing digit images from the inputs, we also generating new digits from randomly sampled latent representations. The synthesized images follow the given class label successfully, showing almost no ambiguity judging with human eyes. Besides, great diversity can be found within the same class.
   To show our model is able to learn the digit representation, examples of interpolation between two different digit images are also provided in Figure~\ref{fig:mnist_inter}. Specifically, we interpolate between the respective latent encodes of the two digits, and the generated digit images show continuous changes. Digit synthesis on the MNIST dataset verifies that our method can be used in conditional image generation, and reflects the superiority of the proposed unified discriminative-generative framework again.

%As illustrated in Figure~\ref{fig:mnist}, the reconstructions of variational capsules are robust while keeping only important details.
%  Apart from reconstructing digit images from the inputs, we also perform linear interpolation between two different digit images to show our model is able to learn the digit representation. Specifically, we interpolate between the respective latent encodes of the two digits. Sample results are shown in Figure~\ref{fig:mnist}.
%
%  As illustrated in Figure~\ref{fig:mnist}, the reconstructions of variational capsules are robust while keeping only important details.
%  Apart from reconstructing digit images from the inputs, we also perform linear interpolation between two different digit images to show our model is able to learn the digit representation. Specifically, we interpolate between the respective latent encodes of the two digits. Sample results are shown in Figure~\ref{fig:mnist}.
%cifar10

%CelebA 91.24%
%Accuracy: 94.6198, 84.0046, 82.5619, 84.5506, 98.9680, 96.0525, 70.5190, 84.7911, 89.7155, 96.0074, 96.1126, 89.2446, 92.9115, 95.8321, 96.5585, 99.6944, 97.6004, 98.3268, 91.8645, 87.7918, 98.4020, 93.9986, 96.8891, 86.7198, 96.3080, 75.8942, 96.8340, 77.2668, 93.5878, 95.2810, 97.8559, 93.0518, 82.8374, 82.5919, 90.6572, 99.0382, 94.1839, 86.2088, 96.1878, 88.0924, Average Accuracy: 91.2404

\section{Conclusion}
In this paper, we have presented a new type of capsule that models images in a unified discriminative and generative framework. The proposed variational capsules are designed in a probabilistic way, in which the values of active capsules are expected to be drawn from a known prior distribution. Thus, the divergence of each capsule with the prior distribution can be used to represent the presence of an entity, deriving a new metric for image classification. By sampling values for active capsules from the prior distribution, the proposed VCs can be further extended into a generative model and employed to synthesize new images. Benefitting from the semantically-disentangling representations learned via VCs, it is easy to synthesize image samples with fine-grained semantic attributes and large diversity. The experimental results of the digit recognition and synthesis as well as the facial attribute prediction and manipulation demonstrate our method's superiority in integrating image analysis and synthesis into a unified framework.

\small
\bibliographystyle{icml2017}
\bibliography{vcbib}

%\subsubsection*{Acknowledgments}
%
%Use unnumbered third level headings for the acknowledgments. All
%acknowledgments go at the end of the paper. Do not include
%acknowledgments in the anonymized submission, only in the final paper.

%\section*{References}
%
%References follow the acknowledgments. Use unnumbered first-level
%heading for the references. Any choice of citation style is acceptable
%as long as you are consistent. It is permissible to reduce the font
%size to \verb+small+ (9 point) when listing the references. {\bf
%  Remember that you can use more than eight pages as long as the
%  additional pages contain \emph{only} cited references.}
%\medskip
%
%\small
%
%[1] Alexander, J.A.\ \& Mozer, M.C.\ (1995) Template-based algorithms
%for connectionist rule extraction. In G.\ Tesauro, D.S.\ Touretzky and
%T.K.\ Leen (eds.), {\it Advances in Neural Information Processing
%  Systems 7}, pp.\ 609--616. Cambridge, MA: MIT Press.
%
%[2] Bower, J.M.\ \& Beeman, D.\ (1995) {\it The Book of GENESIS:
%  Exploring Realistic Neural Models with the GEneral NEural SImulation
%  System.}  New York: TELOS/Springer--Verlag.
%
%[3] Hasselmo, M.E., Schnell, E.\ \& Barkai, E.\ (1995) Dynamics of
%learning and recall at excitatory recurrent synapses and cholinergic
%modulation in rat hippocampal region CA3. {\it Journal of
%  Neuroscience} {\bf 15}(7):5249-5262.

\end{document}